\documentclass[conference]{IEEEtran}
\IEEEoverridecommandlockouts

\usepackage{cite}
\usepackage{amsmath,amssymb,amsfonts}
\usepackage{algorithmic}
\usepackage{graphicx}
\usepackage{textcomp}
\usepackage{xcolor}
\usepackage{multirow}
\usepackage{footmisc}
\usepackage{comment}
\usepackage{amsmath}

\def\BibTeX{{\rm B\kern-.05em{\sc i\kern-.025em b}\kern-.08em
    T\kern-.1667em\lower.7ex\hbox{E}\kern-.125emX}}

\begin{document}

\title{FeatureCuts: Feature Selection for Large Data by Optimizing the Cutoff\\
}

\makeatletter
\newcommand{\linebreakand}{%
  \end{@IEEEauthorhalign}
  \hfill\mbox{}\par
  \mbox{}\hfill\begin{@IEEEauthorhalign}
}
\makeatother

\author{\IEEEauthorblockN{Andy Hu}
\IEEEauthorblockA{
\textit{Commonwealth Bank of Australia}\\
Sydney, Australia \\
andy.hu@cba.com.au}
\and
\IEEEauthorblockN{Devika Prasad}
\IEEEauthorblockA{
\textit{Commonwealth Bank of Australia}\\
Perth, Australia \\
devika.prasad@cba.com.au}
\and
\IEEEauthorblockN{Luiz Pizzato}
\IEEEauthorblockA{
\textit{Commonwealth Bank of Australia}\\
Sydney, Australia \\
luiz.pizzato1@cba.com.au}
\linebreakand
\IEEEauthorblockN{Nicholas Foord}
\IEEEauthorblockA{
\textit{Commonwealth Bank of Australia}\\
Adelaide, Australia \\
nicholas.foord@cba.com.au}
\and
\IEEEauthorblockN{Arman Abrahamyan}
\IEEEauthorblockA{
\textit{Commonwealth Bank of Australia}\\
Sydney, Australia \\
arman.abrahamyan@cba.com.au}
\and
\IEEEauthorblockN{Anna Leontjeva}
\IEEEauthorblockA{
\textit{Commonwealth Bank of Australia}\\
Sydney, Australia \\
anna.leontjeva@cba.com.au}
\linebreakand
\IEEEauthorblockN{Cooper Doyle}
\IEEEauthorblockA{
\textit{Commonwealth Bank of Australia}\\
Sydney, Australia \\
cooper.doyle@cba.com.au}
\and 
\IEEEauthorblockN{Dan Jermyn}
\IEEEauthorblockA{
\textit{Commonwealth Bank of Australia}\\
Sydney, Australia \\
dan.jermyn@cba.com.au}
}

\maketitle

\begin{abstract}
In machine learning, the process of feature selection involves finding a reduced subset of features that captures most of the information required to train an accurate and efficient model. This work presents FeatureCuts, a novel feature selection algorithm that adaptively selects the optimal feature cutoff after performing filter ranking.  Evaluated on 14 publicly available datasets and one industry dataset, FeatureCuts achieved, on average, 15 percentage points more feature reduction and up to 99.6\% less computation time while maintaining model performance, compared to existing state-of-the-art methods. When the selected features are used in a wrapper method such as Particle Swarm Optimization (PSO), it enables 25 percentage points more feature reduction, requires 66\% less computation time, and maintains model performance when compared to PSO alone. The minimal overhead of FeatureCuts makes it scalable for large datasets typically seen in enterprise applications.
\end{abstract}


\section{Introduction}

Traditional machine learning methods work best when their prediction signals come from data with a small, but highly informative set of features. Conversely, when the data contains many features and the prediction signals are sparse across the feature base, these machine learning models become less efficient and can suffer from the curse of dimensionality~\cite{Wang2016}. The curse of dimensionality refers to various phenomena that occur when analyzing or organizing data in high-dimensional space, such as overfitting, computational complexity, increased sparsity, and challenges in interpreting distance metrics~\cite{Li_etal_ACM_2017}. Therefore, it is often beneficial to perform dimensionality reduction, such as feature selection.

Feature selection aims to choose a subset of relevant features from the original features while retaining or improving model performance~\cite{Wang2016}. Unlike other dimensionality reduction techniques that map features into a lower-dimensional new feature space, feature selection maintains the original features, offering better readability and interpretability~\cite{Li_etal_ACM_2017}.

With the widespread adoption of large language models (LLMs), vector embeddings --- often ranging from 768 to 4,096 dimensions --- have become more common in downstream machine learning tasks~\cite{brown2020language}~\cite{bommasani2021opportunities}. In industrial settings, datasets frequently contain hundreds of thousands to millions of instances~\cite{wu2016googles}. As vector embeddings become higher-dimensional and data volumes grow, the applicability and scalability of current feature selection techniques remain underexplored~\cite{Li_etal_ACM_2017}. 

\subsection{Traditional Methods}
Feature selection methods can be categorized into filter, embedded, and wrapper methods~\cite{DBLP:journals/corr/LiCWMTTL16}. Filter methods rank and assess the relevance of features based on statistical measures such as chi-squared test and information gain~\cite{Venkatesh2019326}, which are effective in ranking features independently of any learning algorithm~\cite{Mwadulo_IJCATR_2016}. This independence allows filter methods to be fast and scalable. However, they cannot capture feature dependencies and interactions with the learning algorithm, which can lead to suboptimal model performance~\cite{Ibrahim_etal_PJST_2018}. 

Embedded methods integrate feature selection directly into the model training process,
which can improve the results while maintaining efficiency, such as through regularization
techniques~\cite{Mwadulo_IJCATR_2016}. Embedded methods strike a balance in terms of computational
costs, overfitting and the ability to capture feature dependencies and interactions with the
learning algorithm. However, they require in-depth knowledge of the model parameters and
are tied to specific algorithms~\cite{Liu_etal_IEEEAccess_2018}.

Wrapper methods evaluate feature subsets based on their performance with a specific
learning algorithm~\cite{Mwadulo_IJCATR_2016} which allows them to consider feature interactions and
dependencies more effectively than filter methods. However, this approach is more
computationally intensive and may lead to overfitting.

\subsection{Recent Methods}
Recent wrapper methods have primarily focused on evolutionary feature selection (EFS) algorithms 
which are inspired by biological evolution, such as Particle Swarm Optimization (PSO). EFS algorithms employ a heuristic search approach
that enables a thorough exploration of the feature space, often outperforming traditional wrapper methods in identifying global optima~\cite{SONG2024101661}.
However, despite these advantages, EFS algorithms can be computationally expensive due to the reliance on training
and evaluating the learning algorithm for each feature subset. The substantial time consumption becomes particularly problematic when applied to 
large-scale datasets typical in enterprise environments~\cite{Feng_PLOSONE_2024}.

The computational costs associated with EFS algorithms prompted the development of
hybrid feature selection methods which begin with a filter approach to rapidly reduce the
feature search space before the application of EFS to capture more complex feature
interactions. These methods have been shown to effectively reduce features and improve
classification accuracy while managing computational efficiency~\cite{Stefano_etal_2017, pr11020562}. 

\subsection{Challenge in Finding the Best Feature Cutoff}
The first stage of hybrid feature selection involves sorting the features by a filter method and selecting a
subset of those features to input into an EFS algorithm. A challenge remains in determining
the optimal number of features to select for this subset. There is currently no standardized
method for this critical decision-making step~\cite{Rajab_Wang_JIKM:2020, Ali_et_al_PLOSONE_2018}.

Some methods use a fixed cutoff, such as selecting the top 5 percent of features~\cite{pr11020562}, while others use the mean filter score~\cite{Cateni_et_al_IEEE_2014} or test a
range of arbitrary feature numbers such as the top 100 or top 500 features~\cite{Stefano_etal_2017}. Using a fixed cutoff can lead to suboptimal model
performance if not addressed properly~\cite{Tawhid_DeSouza_ACI_2018} as it might not adapt to
different datasets and result in the exclusion of important features. On the other hand,
iterating over many arbitrary feature ranges is computationally inefficient. This work
addresses the problem of finding the optimal filter feature selection cutoff point.

\subsection{Proposed Approach}
The main contribution of this work is the introduction of a novel methodology named FeatureCuts to automatically select the
feature cutoff after filter-based feature ranking. We reformulate the selection process as an optimization problem 
and propose a Bayesian Optimization and Golden Section Search framework that adaptively selects the optimal cutoff with minimal overhead.
We demonstrate this approach is suitable to be embedded in hybrid feature selection and achieves superior results and reduced computation time across large-scale datasets and with features constructed from LLM vector embeddings.

The remainder of the paper describes FeatureCuts and its algorithm in Section 2, our experiment setup in Section 3, the validation of our filter metric and optimization of the cutoff in Section 4, evaluation of our method in Section 5 and concluding remarks in Section 6.

\begin{figure*}[ht]
  \centering
  \includegraphics[width=1\textwidth, trim={4cm 4cm 4cm 4cm}, clip]{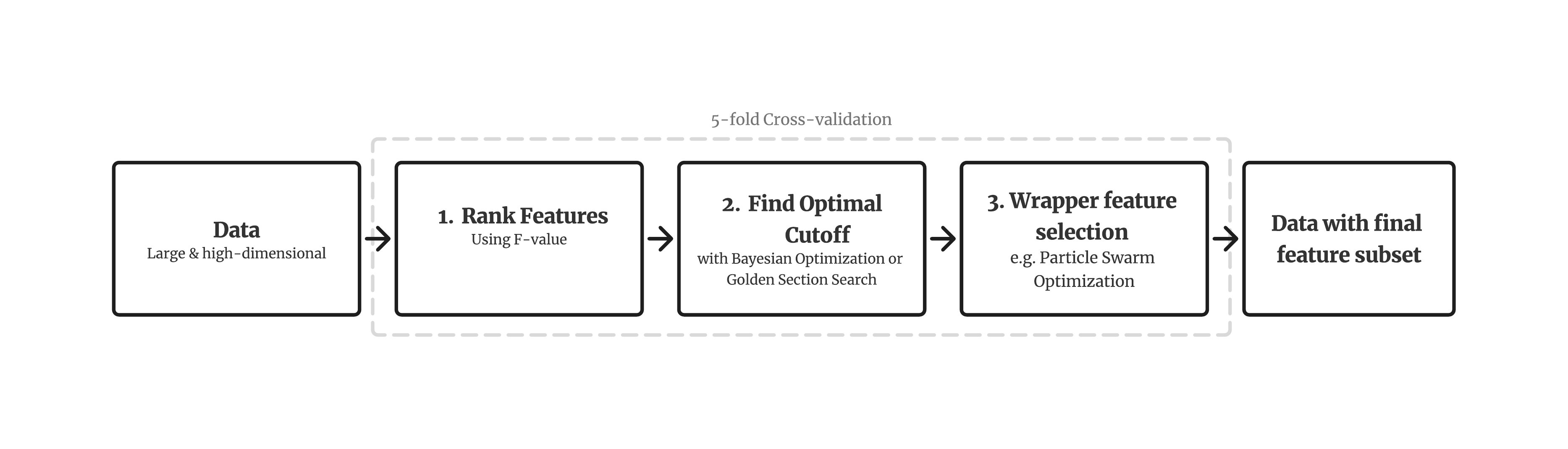}
  \caption{Overview of the FeatureCuts methodology showing the three stages: feature ranking, finding the optimal cutoff, and final feature selection with PSO.}
  \label{f:featurecuts_diagram}
\end{figure*}

\section{FeatureCuts Methodology}

Our proposed hybrid feature selection method consists of three stages as shown in Figure~\ref{f:featurecuts_diagram}: (1) ranking features by performing an F-test between
the features and modeling target variable to prioritize important variables; (2) applying an adaptive filtering method to find the
optimal cutoff point and narrow the feature space; and (3) final selection with Particle Swarm Optimization that selects
a smaller set of features while maintaining model performance.

\subsection{Rank Features}
From the original dataset with many features, all features are individually ranked by a filter
feature selection metric. We used the ANOVA F-value for classification and F-statistic for
regression tasks and both are referred to as the F-value. While any filtering metric can be
used, Section~\ref{s:filter_validation} expands on how we selected F-value over other popular filtering metrics.

\subsection{Evaluating Feature Selection Cutoff}
The objective of feature selection is to obtain a reduced set of features that improves or at
least does not significantly decrease the performance of the models built upon it. As we are
maximizing both feature set reduction and performance, we use the weighted harmonic
mean of the percentage of features removed and the model test score after feature selection. We assign weights of 1 and 50 to feature reduction and model performance respectively. We observed that using these weights
achieved a good balance between feature reduction and model performance on our
evaluation datasets. These weights are similar to previous literature where model
performance is weighted more heavily than feature reduction percentage~\cite{a14030100, PENG2023261}. We call this a feature selection score (FS-score)
and apply this score to evaluate any cutoff after feature ranking and to evaluate the final set
of features selected. The FS-score is mathematically represented as follows:

Let:
\begin{itemize}
  \item $S$ = model score after feature selection
  \item $F_r$ = number of features removed
  \item $F_b$ = original number of features
  \item $w_s = 50$ = weight for model performance
  \item $w_f = 1$ = weight for feature reduction
\end{itemize}

Then the FS-score is defined as:

\[
\text{FS-score} = \frac{w_s + w_f}{\frac{w_s}{S} + \frac{w_f}{1 - \frac{F_r}{F_b}}}
\]

\subsection{Finding the Optimal Cutoff}
After ranking features by their F-value, we need to decide a cutoff feature number to input a
subset of features for the next stage of FeatureCuts. The cutoff that returns the maximum
FS-score is considered as the optimal cutoff. 

We define the following notation:
\begin{itemize}
    \item $N$ be the total number of features.
    \item $k$ denotes the cutoff feature number, where $k \in \{1, 2, \ldots, N\}$.
    \item $FSS(k)$ denotes the FS-score obtained by selecting the top $k$ features.
\end{itemize}

Our objective is to find the optimal cutoff $k^*$ that yields the maximum FS-score. This can be mathematically represented as:

\[
k^* = \underset{k \in \{1, 2, \ldots, N\}}{\arg\max} \; FSS(k)
\]

The maximum FS-score can be found by brute-force iterating through each cutoff $k \in \{1, 2, \ldots, N\}$. 
However, testing each cutoff requires model training, evaluation and calculation of the FS-score which is not
computationally efficient, especially for large data where $N$ is large. Since evaluating each $k$ is computationally expensive and can be viewed as a black-box function, we use Bayesian Optimization~\cite{bayesopt} to efficiently find the optimal $k^*$ within reasonable computational limits

Alternatively, we achieved similar results in less computation time by utilizing Golden Section Search\footnote{We implemented our own function in Python with reference to~\cite{goldenratio_2022}} to find the optimal $k^*$. Although the search is originally designed for continuous unimodal functions it can be adapted for discrete integer intervals. To find the maximum FS-score we set up the Golden Section Search as follows:

Let:
\begin{itemize}
    \item $N$ be the total number of features.
    \item $k$ denote the cutoff value (number of features selected), where $k \in \{1, 2, \ldots, N\}$.
    \item $FSS(k)$ denote the fitness score as a function of $k$.
    \item $a, b$ denote the endpoints of the current search interval, initialized as $a = 1, b = N$.
    \item $\varphi = \frac{\sqrt{5} - 1}{2}$ be the golden ratio conjugate.
    \item $I$ be the maximum number of iterations. $I = 10$.
\end{itemize}

We run Golden Section Search in the interval $[1, N]$ for $I$ iterations. The final cutoff is selected as:
\[
k^* = \underset{k \in \{\lfloor a \rfloor, \lceil b \rceil\}}{\arg\max}~ FSS(k)
\]

\subsection{Wrapper Algorithm on Reduced Feature Set} 
After ranking features by F-value and finding the optimal feature cutoff $k*$, the top $k*$ features can be selected and used as input to a wrapper feature selection method for further feature reduction. These methods iteratively test different combinations of feature subsets and account for feature interdependencies. We used the Py\_FS package~\cite{Guha2022} and tested four evolutionary computing based wrapper methods below:
\begin{itemize}
    \item Particle Swarm Optimization (PSO)~\cite{Ghosh2019}
    \item Grey Wolf Optimization (GWO)~\cite{Dhargupta2020}
    \item Whale Optimization Algorithm (WOA)~\cite{Guha2020ECWSA}
    \item Sine Cosine Algorithm (SCA)~\cite{Smith2021}
\end{itemize}

\section{Experiment Setup}

To validate the effectiveness of FeatureCuts, we conducted a series of experiments and
evaluated based on feature reduction percentage, model performance and computation
time. We aim to address the following key questions through our experiments:
\begin{itemize}
    \item {How does our method compare with existing state-of-the-art methods?}
     \item {What is the impact of using different filter metrics and optimization techniques?}
    \item {Can our method handle large-scale datasets with high-dimensional features
effectively?}
    \item {Can our method be embedded into hybrid feature selection and reduce the computation time of wrapper algorithms while maintaining model performance?}
\end{itemize}

\subsection{Datasets}

\begin{table}[htbp]
\caption{Datasets used to test the feature selection method. Dataset type is either C for classification or R for regression, reflecting the machine learning task.}
\label{t:datasets-table}
\renewcommand{\arraystretch}{1.5}
\resizebox{\columnwidth}{!}{%
\begin{tabular}{|l|c|r|r|r|c|l|}
\hline
\textbf{Dataset} & \textbf{Type} & \textbf{Features} & \textbf{Instances} & \textbf{Used Samples} & \textbf{Label Classes} & \textbf{Reference} \\
\hline
airline   & C       & 1,920  & 540,000 & 100,000   & 2 &  \cite{kaggleAirlineDataset}   \\
article   & C       & 1,536  & 10,000  & 10,000   & 5  &  \cite{kaggleNewsArticles} \\
blog      & R       & 280    & 60,021  & 19,966   & -  & \cite{blogfeedback_304} \\
bss      & C       & 3,572    & 13,991  & 20,000   & 2  &  Proprietary Dataset\\
carer     & C       & 384    & 16,000  & 16,000   & 6 &   \cite{saravia-etal-2018-carer} \\
colon     & C       & 2,000  & 62      & 62       & 2 &   \cite{colon} \\
cyber     & C       & 384    & 47,692  & 20,000   & 6 &   \cite{9378065} \\
gisette   & C       & 5,000  & 7,000   & 7,000    & 2 &   \cite{gisette_170} \\
house     & R       & 79     & 1,460   & 1,460    & - & \cite{house-prices-advanced-regression-techniques} \\
isolet    & C       & 617    & 7,797    & 7,797     & 26  & \cite{isolet_54} \\
madelon   & C       & 500    & 2,000   & 2,000    & 2    & \cite{madelon_171} \\
mnist     & C       & 784    & 60,000  & 20,000   & 10   & \cite{kaggleMNISTDataset} \\
relathe   & C       & 4,322  & 1,427   & 1,427    & 2    & \cite{Lang95} \\
spam      & C       & 384   & 193,849 & 19,385   & 2    & \cite{kaggle190KSpam} \\
spam-meta & C      & 2,048 & 193,849 & 2,550   & 2    & \cite{kaggle190KSpam} \\
\hline
\end{tabular}
}
\end{table}

We used fourteen openly available datasets with some that are commonly used to evaluate feature
selection methods, sourced mostly from the UCI Machine Learning Repository~\cite{kelly2024uci} and Kaggle~\cite{kaggle}. We included one proprietary dataset to evaluate our method's applicability on industry data. These datasets represent a mix of classification and regression
problems with both binary and multi-class classification. We sampled up to 100,000
instances to represent the scale of big data and up to 5,000 features. The details of these
datasets are described in Table~\ref{t:datasets-table}. Sentence embeddings were generated for all text columns in the datasets using the all-MiniLM-L6-v2 model
except for the spam-meta dataset, where we used the Meta-Llama-3-8B model to obtain higher-dimensional embeddings for evaluation. The sentence embeddings were concatenated and expanded, with each embeddings' dimensions becoming a feature. 

\subsection{Comparison with Other Methods}
We compared FeatureCuts to the evolutionary-based wrapper methods PSO, GWO, WOA and SCA by running them directly on the evaluation datasets. Outside of these evolutionary methods, we also tested Boruta~\cite{kursa2010feature}. To account for the F-score being univariate and the potential limitation of missing feature interactions we also compared with using ReliefF~\cite{Urbanowicz2017Benchmarking} to rank features followed by applying Golden Section Search to find the optimal cutoff. 

\subsection{Cutoff Parameters}

For the Bayesian Optimization parameters we used 5 initial points and 5 iterations to balance exploration and computational efficiency. We experimented with combinations of 5, 20, 50 and 80 initial points and 5 or 10 iterations and found that using 5 initial points and 5 iterations achieved a good balance between performance and computation time.

For Golden Section Search we set the number of iterations to 10.

\subsection{Other Methods' Parameters}

For PSO, GWO, WOA and SCA we used 20 agents and 50 maximum iterations.
For Boruta, we set the maximum depth to 5 and class weight to balanced for the random forest regressor and classifier. The number of estimators was set to `auto' and all other parameters were left as default in the Boruta package.

\subsection{Model and Evaluation Metric}

All evaluations were done using a nested, stratified 5-fold cross-validation and hold-out test
set. For classification tasks we used XGBoost classifier and evaluated with ROC AUC for
binary targets and F1 for multiclass. For regression tasks we used XGBoost regressor
and evaluated with $R^2$.

For baseline comparisons with PSO, GWO, WOA and SCA the fitness function was set so the algorithm optimizes on the FS-score.

When using the features selected from FeatureCuts as input into the wrapper algorithms PSO, GWO, WOA and SCA, the fitness function was set so that the algorithm optimizes on model performance only. This is to maintain model performance after the rapid feature reduction already achieved by FeatureCuts.

\begin{table}[t]
\caption{ROC AUC scores at each feature cutoff $k$ for the airline dataset across each filter ranking metric. The highest scores for each $k$ value are highlighted in bold.}
\begin{center}
\renewcommand{\arraystretch}{1.3}

\begin{tabular}{|c|c|c|c|c|c|c|}
\hline
\textbf{Metric} & \textbf{k=50} & \textbf{k=100} & \textbf{k=200} & \textbf{k=500} & \textbf{k=750} & \textbf{k=1,000} \\
\hline
MIC        & 0.52           & 0.53           & 0.52           & 0.61           & 0.60           & 0.62 \\
MI         & \textbf{0.64}  & \textbf{0.64}  & \textbf{0.64}  & 0.61           & 0.61           & 0.60 \\
F\_classif & \textbf{0.64}  & 0.63           & 0.63           & \textbf{0.64}  & \textbf{0.63}  & \textbf{0.63} \\
Var        & 0.63           & 0.63           & 0.63           & 0.63           & 0.62           & 0.62 \\
Corr       & 0.63           & 0.63           & \textbf{0.64}  & 0.63           & 0.62           & \textbf{0.63} \\
IG         & 0.52           & 0.52           & 0.52           & 0.60           & 0.61           & 0.62 \\
IGR        & 0.53           & 0.53           & 0.52           & 0.58           & 0.59           & 0.61 \\
\hline
\end{tabular}
\label{t:auc_k}
\end{center}
\end{table}

\section{Filter Metric and Cutoff Validation}

\subsection{Choosing the Metric}
\label{s:filter_validation}
To find the most suitable filtering metric, popular metrics such as mutual information (MI), variance (Var),
maximal information coefficient (MIC), information gain (IG), information gain ratio (IGR), correlation (Corr) and chi-squared (Chi2)
were independently tested alongside F-value (f-classif and f-regres). Each evaluation
dataset was ranked by these metrics and a range of cutoffs ($k$), from 50 to 1,000 was tested for the model performance.

For most values of $k$, MI
and F-value had the most datasets with the highest model score. An example for the airline dataset in Table~\ref{t:auc_k} shows ranking features by MI or F-classif returned the highest model performance across all $k$ values.

\begin{figure*}[!ht]
  \begin{center}
  \centerline{\includegraphics[width=1\textwidth]{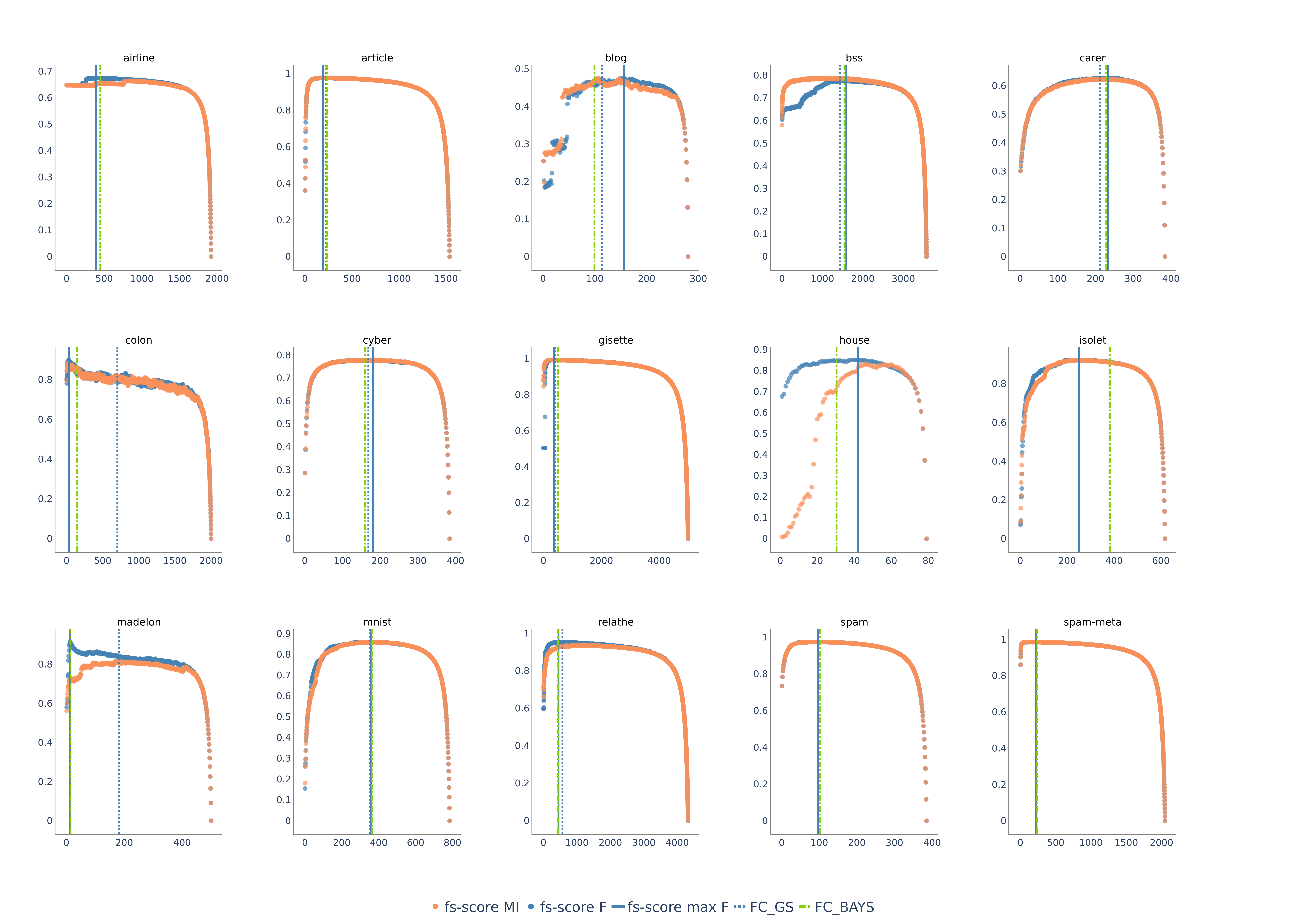}}
  \caption{Plots of the FS-scores at all feature cutoffs after using F-value and MI to rank features across the test datasets. The cutoffs found by Golden Section Search (FC\_GS) and Bayesian Optimization (FC\_BAYS) are displayed in dotted vertical lines. This is compared to the maximum FS-scores based on F-value (fs-score max F) in solid blue lines}
  \label{f:brute_plots}
  \end{center}
\end{figure*}

\subsection{Brute Force Cutoff Comparison}
To validate the performance of our cutoff selection method, we needed a ground truth for what the "optimal" cutoff is. However, for most real-world datasets, no ground truth exists. To address this, we conducted a brute-force search to find the true optimal feature cutoff.

Specifically, we first identified MI and F-value as the best ranking metrics from previous section~\ref{s:filter_validation}. For each dataset, we ranked features by MI and F-value scores. Then, for every possible feature cutoff $k$, from 1 to $N$ features, we trained a model using just the top $k$ features and recorded the resulting FS-score. This exhaustive evaluation for each $k$ enabled us to pinpoint, for each dataset, the cutoff which yielded the highest FS-score. We then treated this value as the "ground truth optimal cutoff" for the purpose of benchmarking our method.

Figure~\ref{f:brute_plots} visualizes this process: FS-score as a function of the cutoff $k$, using MI and F-value for ranking. Notably, across most datasets, F-value based selection (blue curve) generally reaches higher FS-score peaks than MI (orange curve), so we selected F-value as our preferred filter metric for FeatureCuts. We then define the following abbreviations:
\begin{itemize}
    \item FC\_GS - cutoff found by Golden Section Search
    \item FC\_BAYS - cutoff found by Bayesian Optimization
\end{itemize}
The dashed and dotted vertical lines indicate the FC\_GS and FC\_BAYS cutoffs. As shown, these automated selections typically align closely with the brute-force maximum (solid blue line), demonstrating that FeatureCuts can efficiently approximate the empirically optimal feature cutoff.

\section{Evaluation}

To assess our method’s ability to balance feature reduction and model performance in reasonable computation time, we ran the adaptive filtering stage of FeatureCuts (finding the optimal cutoff with Bayesian Optimization and Golden Section Search, then selecting the top $k$ features) on the evaluation datasets. We compared our results to PSO, GWO, SCA, WOA, Boruta and using ReliefF to rank features. We also evaluated the suitability of FeatureCuts to be embedded into hybrid feature selection by inputting the selected top $k$ features into PSO, GWO, SCA and WOA.

\subsection{Comparing FeatureCuts With Other Methods}

The averaged results across all datasets can be found in Table~\ref{t:avg_baseline}. 
Our approach which automatically selects the optimal cutoff, requires significantly less computation time than all the competing methods, with an average run time of 1 minute and 3 seconds. This is a substantial improvement over the other methods, which took approximately 19 minutes for Boruta and 3-4 hours for PSO, GWO, SCA and WOA. FeatureCuts also achieved the highest feature reduction percentage while maintaining a competitive test score.

\begin{table}[!thbp]
\centering
\caption{Comparison of FeatureCuts to Other Methods, Averaged With Std. Deviation Across All Datasets}
\renewcommand{\arraystretch}{1.3}
\resizebox{\columnwidth}{!}{%
\begin{tabular}{|l|c|c|c|}
\hline
\textbf{Method} & \textbf{\% Feats Reduction} & \textbf{Test Score} & \textbf{Total Time (hh:mm:ss)} \\
\hline
No FS     & 0.00 $\pm$ 0.00             & 0.823 $\pm$ 0.17      & 00:00:00 $\pm$ 00:00:00 \\
FC\_GS    & 67.25 $\pm$ 16.32           & 0.817 $\pm$ 0.17      & \textbf{00:01:03} $\pm$ 00:01:21 \\
FC\_BAYS  & \textbf{67.99} $\pm$ 16.88           & 0.820 $\pm$ 0.17      & 00:01:12 $\pm$ 00:01:17 \\
Boruta    & 56.65 $\pm$ 35.62           & \textbf{0.825} $\pm$ 0.18      & 00:19:02 $\pm$ 00:16:03 \\
ReliefF   & 65.59 $\pm$ 19.45           & 0.822 $\pm$ 0.17      & 00:46:35 $\pm$ 01:07:15 \\
PSO       & 57.48 $\pm$ ~5.27            & 0.819 $\pm$ 0.18      & 03:04:04 $\pm$ 03:14:09 \\
GWO       & 52.77 $\pm$ 12.95           & 0.814 $\pm$ 0.17      & 03:59:42 $\pm$ 04:12:07 \\
SCA       & 53.43 $\pm$ 12.37           & 0.819 $\pm$ 0.17      & 04:09:42 $\pm$ 04:26:52 \\
WOA       & 57.71 $\pm$ 13.25           & 0.821 $\pm$ 0.17      & 03:51:47 $\pm$ 04:05:39 \\
\hline
\end{tabular}
\label{t:avg_baseline}
}
\end{table}

The distribution of feature reduction percentages and computation times across all datasets is shown in Figure~\ref{f:feat_reduct}, and test scores in Table~\ref{t:test_scores}. When analyzing the model scores across individual datasets, Boruta achieved uniquely the best model score on 5 out of the 15 datasets, however Boruta is limited to tree-based models. Additionally, Boruta had the worst consistency in performing feature reduction with a reduction percentage below 45 for many datasets. 
When compared to using ReliefF to rank features before applying Golden Section Search to find the cutoff, ReliefF achieved better model performance on 4 datasets, worse on 2 and performed similarly across the rest. This could be explained by how ReliefF captures feature interactions when ranking features. However FeatureCuts still achieved similar feature reduction across all datasets and is on average faster.

FeatureCuts performed notably well on datasets containing text examples, with features corresponding to sentence embedding dimensions computed using an LLM. This is shown in Table~\ref{t:test_scores}. On these datasets marked by a dagger($\dagger$), the test score was the same or 0.01 lower than the best score achieved across all methods.

Overall the results demonstrate that FeatureCuts achieved substantial feature reduction with consistency while maintaining model performance in significantly less computation time compared to other approaches. The full results table showing feature reduction, test score and run time for each dataset and method are shown in Table~\ref{f:full_results_comparison} in the Appendix.

\begin{table}[!thbp]
\caption{Test Score comparison between FeatureCuts and other methods on all datasets.}
\begin{center}
\renewcommand{\arraystretch}{1.3}
\resizebox{\columnwidth}{!}{%
\begin{tabular}{|l|c|c|c|c|c|c|c|c|}
\hline
\textbf{Dataset}
& \textbf{FC\_GS}
& \textbf{FC\_BAYS}
& \textbf{Boruta}
& \textbf{ReliefF}
& \textbf{PSO}
& \textbf{WOA}
& \textbf{GWO}
& \textbf{SCA} \\
\hline
airline$^\dagger$      & \textbf{0.68} & \textbf{0.68} & \textbf{0.68} & 0.67 & \textbf{0.68} & \textbf{0.68} & \textbf{0.68} & \textbf{0.68} \\
article$^\dagger$      & \textbf{0.98} & \textbf{0.98} & 0.94 & \textbf{0.98} & \textbf{0.98} & \textbf{0.98} & \textbf{0.98} & \textbf{0.98} \\
blog                   & 0.37 & 0.39 & 0.33 & 0.40 & 0.37 & \textbf{0.41} & 0.39 & 0.38 \\
bss$^\dagger$          & 0.79 & 0.79 & \textbf{0.80} & \textbf{0.80} & 0.79 & 0.78 & 0.79 & 0.79 \\
carer$^\dagger$        & \textbf{0.64} & \textbf{0.64} & \textbf{0.64} & 0.63 & 0.62 & 0.62 & 0.63 & 0.63 \\
colon                  & 0.88 & 0.88 & \textbf{0.93} & 0.86 & 0.88 & 0.86 & 0.80 & 0.85 \\
cyber$^\dagger$        & 0.79 & 0.79 & \textbf{0.80} & 0.79 & 0.79 & 0.78 & 0.79 & 0.79 \\
gisette                & \textbf{1.00} & \textbf{1.00} & \textbf{1.00} & \textbf{1.00} & \textbf{1.00} & \textbf{1.00} & \textbf{1.00} & \textbf{1.00} \\
house                  & 0.82 & 0.83 & 0.82 & \textbf{0.85} & 0.84 & 0.81 & 0.82 & 0.84 \\
isolet                 & 0.66 & 0.66 & \textbf{0.67} & 0.66 & 0.66 & 0.65 & \textbf{0.67} & \textbf{0.67} \\
madelon                & 0.85 & 0.85 & \textbf{0.93} & 0.91 & 0.87 & 0.87 & 0.86 & 0.87 \\
mnist                  & 0.87 & 0.88 & \textbf{0.89} & 0.87 & 0.88 & 0.88 & 0.88 & 0.88 \\
relathe                & 0.95 & \textbf{0.96} & 0.93 & 0.95 & 0.95 & 0.94 & 0.95 & 0.95 \\
spam$^\dagger$         & 0.98 & 0.98 & \textbf{0.99} & 0.98 & 0.98 & 0.98 & 0.98 & 0.98 \\
spam-meta$^\dagger$    & \textbf{0.99} & \textbf{0.99} & \textbf{0.99} & \textbf{0.99} & \textbf{0.99} & \textbf{0.98} & \textbf{0.99} & \textbf{0.99} \\
\hline
\end{tabular}
}
\\[2mm]
\footnotesize{$^\dagger$Datasets containing text samples, with features corresponding to sentence embedding dimensions computed using an LLM.}
\label{t:test_scores}
\end{center}
\end{table}

\subsection{Suitability of FeatureCuts for Hybrid Feature Selection}

\begin{figure}[ht]
  \centering
  \includegraphics[width=1\columnwidth]{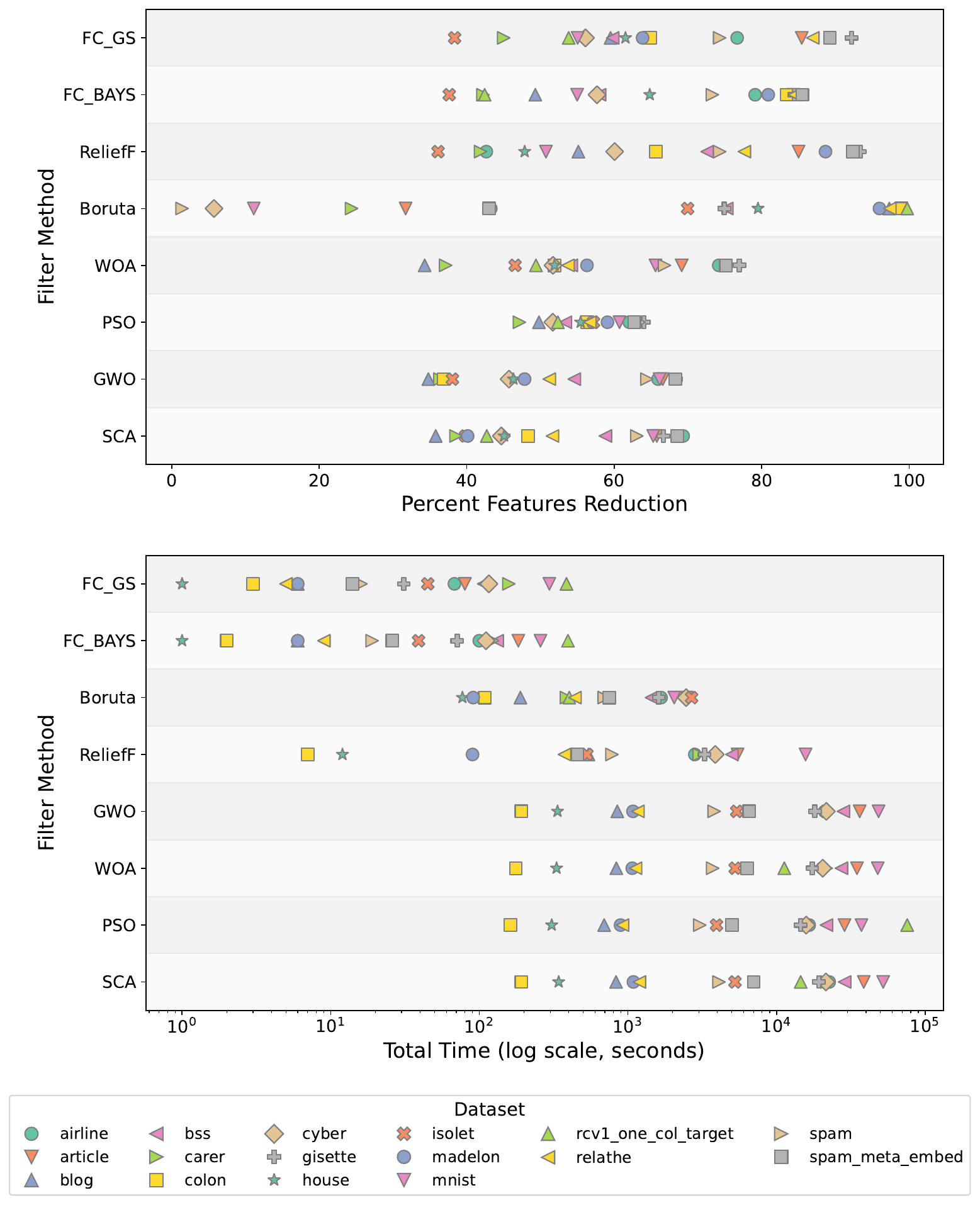}
  \caption{Feature reduction percentages and computation time across different methods. FeatureCuts consistently achieves higher feature reduction in less computation time.}
  \label{f:feat_reduct}
\end{figure}

The average feature reduction percentages, model performances and run time of FeatureCuts when used as a filter ranking step before PSO, GWO, WOA and SCA are shown in Table~\ref{t:avg_hybrid}. The results show that when FeatureCuts is used as a filtering step before PSO, the model performance is maintained from 0.82 to 0.81 while feature reduction percentage improved from 57 to 82 and computation time reduced from 3 hours to 1 hour. WOA achieved similar performance but reduced a lower percentage of features than PSO. For GWO and SCA, model performance drops by 5 percentage points when used as a subsequent wrapper to FeatureCuts, while still benefiting from improved feature reduction and reduced computation time. The consistency of feature reduction is improved as well in comparison to using the cutoff alone, which is shown in Figure~\ref{f:feat_reduct_hybrid} in the Appendix. The full results table for hybrid feature selection is shown in Table~\ref{t:full_results_comparison_hybrid} in the Appendix.

\begin{table}[thbp]
\caption{Suitability of using FeatureCuts as the filtering stage in hybrid feature selection. The best hybrid reuslts are highlighted in bold.}
\centering
\renewcommand{\arraystretch}{1.3}
\resizebox{\columnwidth}{!}{
\begin{tabular}{|l|c|c|c|}
\hline
\textbf{Method} & \textbf{\% Feats Reduction} & \textbf{Test Score} & \textbf{Total Time (hh:mm:ss)} \\
\hline
No FS             & 0.00 $\pm$ 0.00       & 0.823 $\pm$ 0.17     & 00:00:00 $\pm$ 00:00:00 \\
FC\_GS            & 67.25 $\pm$ 16.32     & 0.803 $\pm$ 0.18     & 00:01:04 $\pm$ 00:01:21 \\
\hline
PSO               & 57.48 $\pm$ ~5.27      & 0.819 $\pm$ 0.18     & 03:04:05 $\pm$ 03:14:09 \\
FC\_GS$\rightarrow$PSO & \textbf{82.57 $\pm$ 9.03} & \textbf{0.814 $\pm$ 0.18} & \textbf{01:05:39 $\pm$ 01:34:43} \\
\hline
WOA               & 57.71 $\pm$ 13.25     & 0.821 $\pm$ 0.17     & 03:51:47 $\pm$ 04:05:39 \\
FC\_GS$\rightarrow$WOA & 76.88 $\pm$ 12.61 & 0.812 $\pm$ 0.17 & 01:21:14 $\pm$ 01:55:10 \\
\hline
GWO               & 52.77 $\pm$ 12.95     & 0.814 $\pm$ 0.17     & 03:59:42 $\pm$ 04:12:08 \\
FC\_GS$\rightarrow$GWO & 78.03 $\pm$ 11.10 & 0.758 $\pm$ 0.19     & 01:25:06 $\pm$ 02:02:04 \\
\hline
SCA               & 53.43 $\pm$ 12.37     & 0.819 $\pm$ 0.17     & 04:09:43 $\pm$ 04:26:52 \\
FC\_GS$\rightarrow$SCA & 78.63 $\pm$ 10.75 & 0.765 $\pm$ 0.18     & 01:17:06 $\pm$ 01:49:57 \\
\hline
\end{tabular}
}
\label{t:avg_hybrid}
\end{table}

\section{Conclusion and Future Works}
\label{s:conclusion}

Current feature selection approaches on large and high-dimensional data have
challenges with excessive computation time and balancing feature reduction with model performance. Some
approaches to address these challenges are hybrid filter-based evolutionary feature selection algorithms. However,
these approaches still have limitations in the filtering stage when deciding the number of
features to select and often use a fixed cutoff. In response, FeatureCuts introduces a method to automatically
find a cutoff after filter feature ranking that is close to the optimal cutoff on the FS-score.
Compared to other state-of-the-art feature selection methods, selecting features using the cutoff optimization alone achieves a competitive balance between model performance and feature reduction while requiring
significantly less time. For scenarios requiring further feature reduction, our cutoff optimization method can be integrated into hybrid filter-based feature selection algorithms. Our results show that this approach is particularly effective with Particle Swarm Optimization and on datasets containing transformer-based language vector embeddings. It may also be extendable to other metrics and wrapper feature selection methods which best suit the dataset it is being applied to. Further experimentation could include validation with other machine learning models, other model evaluation metrics and comparison with using ReliefF in hybrid feature selection.

\section*{Acknowledgment}

The authors acknowledge the use of large language models (LLMs) to assist in editing and improving this paper.

\bibliographystyle{IEEEtran}
\bibliography{featurecuts}

\clearpage
\appendix

\section{Suitability For Hybrid Method Full Results}
\begin{figure}[ht]
  \centering
  \includegraphics[width=1\columnwidth]{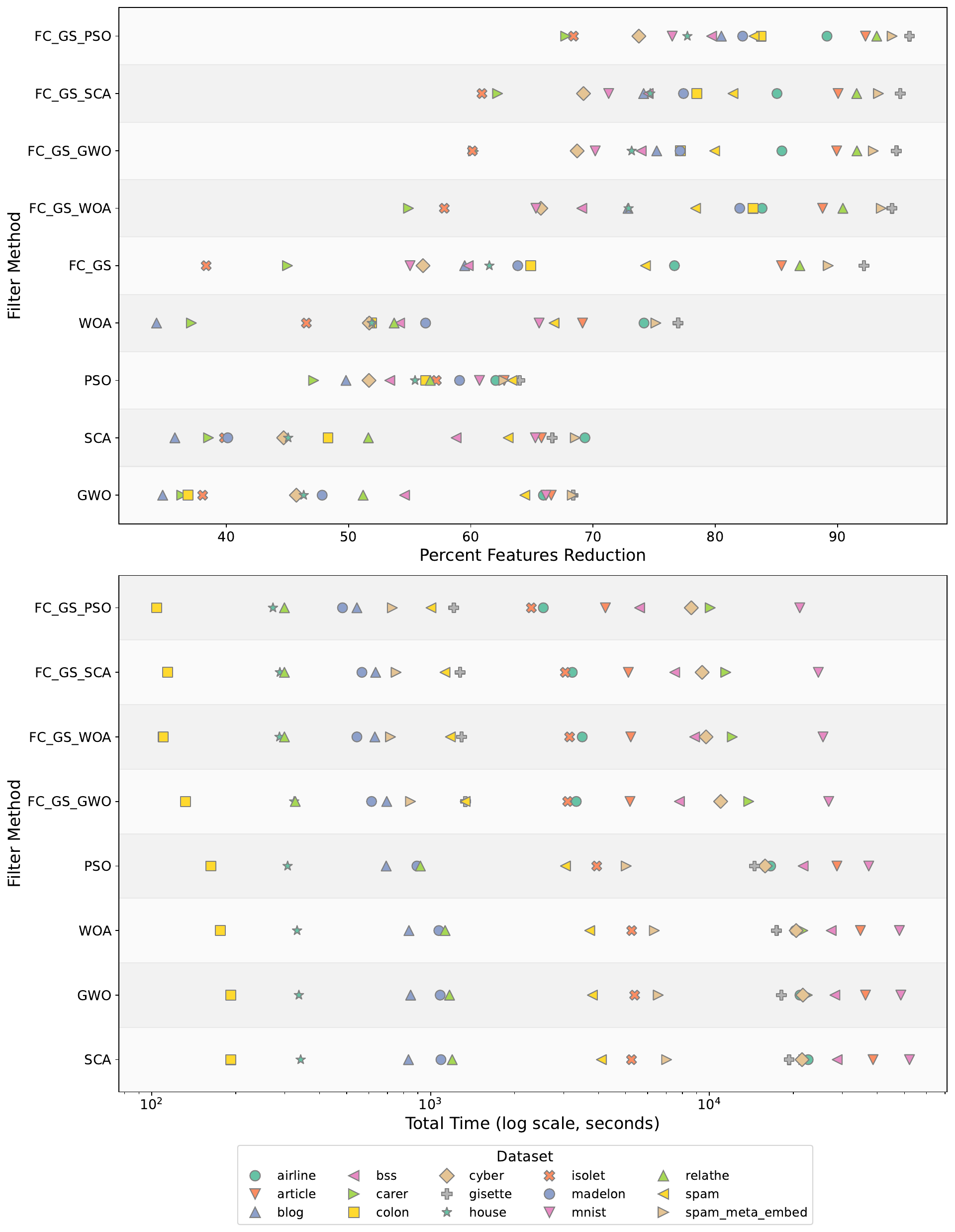}
  \caption{Feature reduction percentages and total computation time when using FeatureCuts before wrapper feature selection methods. FeatureCuts is most suitable as a filter before Particle Swarm Optimization compared to other approaches, achieving the best feature reduction in the shortest time.}
  \label{f:feat_reduct_hybrid}
\end{figure}

\begin{table*}[htbp]
\caption{Full Results Table Comparison With Other Methods. Time is in hh:mm:ss}
\begin{center}
\renewcommand{\arraystretch}{1.3}
\resizebox{\textwidth}{!}{%
\begin{tabular}{|l|l|c|c|c|c|c|c|c|c|c|}
\hline
\textbf{Dataset} & \textbf{Metric} & \textbf{FC\_GS} & \textbf{FC\_BAYS}
& \textbf{ReliefF} & \textbf{PSO} & \textbf{WOA} & \textbf{GWO} & \textbf{SCA}
& \textbf{Boruta} \\
\hline
\multirow{3}{*}{airline$^\dagger$}
& \% Reduction    & 76.66 & \textbf{79.1} & 42.65 & 62.04 & 74.17 & 65.94 & 69.34 & 43.26 \\
& Test Score      & \textbf{0.68} & \textbf{0.68} & 0.67 & \textbf{0.68} & \textbf{0.68} & \textbf{0.68} & \textbf{0.68} & \textbf{0.68} \\
& Time            & \textbf{00:01:08} & 00:01:40 & 00:46:59 & 04:36:18 & 05:38:09 & 05:51:54 & 06:16:43 & 00:27:51 \\
\hline
\multirow{3}{*}{article$^\dagger$}
& \% Reduction    & \textbf{85.42} & 84.3 & 84.99 & 62.73 & 69.14 & 66.58 & 65.78 & 31.71 \\
& Test Score      & \textbf{0.98} & \textbf{0.98} & \textbf{0.98} & \textbf{0.98} & \textbf{0.98} & \textbf{0.98} & \textbf{0.98} & 0.94 \\
& Time            & \textbf{00:01:20} & 00:03:03 & 01:31:08 & 07:57:40 & 09:39:51 & 10:04:43 & 10:43:39 & 00:41:28 \\
\hline
\multirow{3}{*}{blog}
& \% Reduction    & 59.5 & 49.29 & 55.14 & 49.79 & 34.29 & 34.79 & 35.79 & \textbf{97.29} \\
& Test Score      & 0.37 & 0.39 & 0.40 & 0.37 & \textbf{0.41} & 0.39 & 0.38 & 0.33 \\
& Time            & \textbf{00:00:06} & \textbf{00:00:06} & 00:09:02 & 00:11:32 & 00:13:56 & 00:14:08 & 00:13:52 & 00:03:09 \\
\hline
\multirow{3}{*}{bss$^\dagger$}
& \% Reduction    & 59.79 & 58.02 & 72.6 & 53.37 & 54.18 & 54.58 & 58.8 & \textbf{75.26} \\
& Test Score      & 0.79 & 0.79 & \textbf{0.80} & 0.79 & 0.78 & 0.79 & 0.79 & \textbf{0.80} \\
& Time            & \textbf{00:01:47} & 00:02:12 & 01:23:26 & 06:01:16 & 07:35:38 & 07:49:42 & 07:58:36 & 00:24:02 \\
\hline
\multirow{3}{*}{carer$^\dagger$}
& \% Reduction    & 45.0 & 42.19 & 41.87 & \textbf{47.14} & 37.14 & 36.35 & 38.54 & 24.38 \\
& Test Score      & \textbf{0.64} & \textbf{0.64} & 0.63 & 0.62 & 0.62 & 0.63 & 0.63 & \textbf{0.64} \\
& Time            & 00:02:38 & \textbf{00:02:03} & 00:50:18 & 04:29:02 & 06:00:23 & 06:14:04 & 06:03:29 & 00:06:26 \\
\hline
\multirow{3}{*}{colon}
& \% Reduction    & 64.89 & 83.41 & 65.62 & 56.3 & 51.87 & 36.86 & 48.32 & \textbf{98.9} \\
& Test Score      & 0.88 & 0.88 & 0.86 & 0.88 & 0.86 & 0.80 & 0.85 & \textbf{0.93} \\
& Time            & 00:00:03 & \textbf{00:00:02} & 00:00:07 & 00:02:42 & 00:02:56 & 00:03:12 & 00:03:12 & 00:01:49 \\
\hline
\multirow{3}{*}{cyber$^\dagger$}
& \% Reduction    & 56.09 & 57.66 & \textbf{60.05} & 51.67 & 51.69 & 45.73 & 44.69 & 5.73 \\
& Test Score      & 0.79 & 0.79 & 0.79 & 0.79 & 0.78 & 0.79 & 0.79 & \textbf{0.80} \\
& Time            & 00:01:56 & \textbf{00:01:51} & 01:04:28 & 04:23:53 & 05:41:04 & 06:00:34 & 05:58:05 & 00:41:05 \\
\hline
\multirow{3}{*}{gisette}
& \% Reduction    & 92.17 & 84.41 & \textbf{93.32} & 64.01 & 77.14 & 68.38 & 66.66 & 74.92 \\
& Test Score      & \textbf{1.00} & \textbf{1.00} & \textbf{1.00} & \textbf{1.00} & \textbf{1.00} & \textbf{1.00} & \textbf{1.00} & \textbf{1.00} \\
& Time            & \textbf{00:00:31} & 00:01:11 & 00:54:39 & 04:02:01 & 04:47:04 & 05:01:57 & 05:21:42 & 00:26:53 \\
\hline
\multirow{3}{*}{house}
& \% Reduction    & 61.52 & 64.81 & 47.85 & 55.44 & 51.90 & 46.33 & 45.06 & \textbf{79.49} \\
& Test Score      & 0.82 & 0.83 & \textbf{0.85} & 0.84 & 0.81 & 0.82 & 0.84 & 0.82 \\
& Time            & \textbf{00:00:01} & \textbf{00:00:01} & 00:00:12 & 00:05:07 & 00:05:31 & 00:05:36 & 00:05:42 & 00:01:17 \\
\hline
\multirow{3}{*}{isolet}
& \% Reduction    & 38.35 & 37.63 & 36.11 & \textbf{57.18} & 46.55 & 38.06 & 39.84 & 69.95 \\
& Test Score      & 0.66 & 0.66 & 0.66 & 0.66 & \textbf{0.67} & \textbf{0.67} & \textbf{0.67} & \textbf{0.67} \\
& Time            & 00:00:45 & \textbf{00:00:39} & 00:08:50 & 01:05:42 & 01:27:42 & 01:29:52 & 01:27:33 & 00:44:43 \\
\hline
\multirow{3}{*}{madelon}
& \% Reduction    & 63.84 & 80.88 & 88.64 & 59.08 & 56.30 & 47.84 & 40.12 & \textbf{95.96} \\
& Test Score      & 0.85 & 0.85 & 0.91 & 0.87 & 0.87 & 0.86 & 0.87 & \textbf{0.93} \\
& Time            & \textbf{00:00:06} & \textbf{00:00:06} & 00:01:30 & 00:14:52 & 00:17:50 & 00:18:02 & 00:18:09 & 00:01:31 \\
\hline
\multirow{3}{*}{mnist}
& \% Reduction    & 55.03 & 54.97 & 50.74 & 60.71 & \textbf{66.15} & 65.28 & 65.28 & 11.12 \\
& Test Score      & 0.87 & 0.88 & 0.87 & 0.88 & 0.88 & 0.88 & 0.88 & \textbf{0.89} \\
& Time            & 00:04:56 & \textbf{00:04:18} & 04:21:21 & 10:21:09 & 13:20:12 & 13:29:40 & 14:29:39 & 00:34:09 \\
\hline
\multirow{3}{*}{relathe}
& \% Reduction    & 86.92 & 84.43 & 77.64 & 56.68 & 53.73 & 51.19 & 51.62 & \textbf{97.36} \\
& Test Score      & 0.95 & \textbf{0.96} & 0.95 & 0.95 & 0.94 & 0.95 & 0.95 & 0.93 \\
& Time            & \textbf{00:00:05} & 00:00:09 & 00:06:14 & 00:15:19 & 00:18:46 & 00:19:28 & 00:19:55 & 00:07:20 \\
\hline
\multirow{3}{*}{spam$^\dagger$}
& \% Reduction    & \textbf{74.29} & 73.3 & \textbf{74.29} & 63.38 & 66.81 & 64.42 & 63.06 & 1.40 \\
& Test Score      & 0.98 & 0.98 & 0.98 & 0.98 & 0.98 & 0.98 & 0.98 & \textbf{0.99} \\
& Time            & 00:00:16 & 00:00:19 & 00:13:01 & 00:50:45 & 01:01:59 & 01:03:23 & 01:08:17 & \textbf{00:11:32} \\
\hline
\multirow{3}{*}{spam\_meta\_embed$^\dagger$}
& \% Reduction    & 89.24 & 85.49 & \textbf{92.35} & 62.70 & 75.14 & 68.30 & 68.55 & 43.05 \\
& Test Score      & \textbf{0.99} & \textbf{0.99} & \textbf{0.99} & \textbf{0.99} & 0.98 & \textbf{0.99} & \textbf{0.99} & \textbf{0.99} \\
& Time            & \textbf{00:00:14} & 00:00:26 & 00:07:38 & 01:23:49 & 01:45:42 & 01:49:12 & 01:56:59 & 00:12:28 \\
\hline
\end{tabular}
}
\\[1.2ex]
\footnotesize{$^\dagger$ Datasets containing text samples, with features corresponding to sentence embedding dimensions computed using an LLM.}
\end{center}
\label{f:full_results_comparison}
\end{table*}

\begin{table*}[htbp]
\caption{Full Results Table for Suitability For Hybrid Methods. Time is in hh:mm:ss}
\begin{center}
\renewcommand{\arraystretch}{1.4}
\resizebox{\textwidth}{!}{%
\begin{tabular}{|l|l|c|c|c|c|c|c|c|c|}
\hline
\textbf{Dataset} & \textbf{Metric}
& \textbf{PSO} 
& \textbf{FC\_GS$\rightarrow$PSO}
& \textbf{WOA}
& \textbf{FC\_GS$\rightarrow$WOA}
& \textbf{GWO}
& \textbf{FC\_GS$\rightarrow$GWO}
& \textbf{SCA}
& \textbf{FC\_GS$\rightarrow$SCA} \\
\hline
\multirow{3}{*}{airline$^\dagger$}
 & \% Reduction & 62.04 & \textbf{89.14} & 74.17 & 83.83 & 65.94 & 85.45 & 69.34 & 85.05 \\
 & Test Score & \textbf{0.68} & \textbf{0.68} & \textbf{0.68} & \textbf{0.68} & \textbf{0.68} & 0.66 & \textbf{0.68} & 0.66 \\
 & Time & 4:36:18 & \textbf{0:42:16} & 5:38:10 & 0:58:18 & 5:51:54 & 0:55:30 & 6:16:43 & 0:53:42 \\
\hline
\multirow{3}{*}{article$^\dagger$}
 & \% Reduction & 62.73 & \textbf{92.29} & 69.14 & 88.78 & 66.58 & 89.93 & 65.78 & 90.05 \\
 & Test Score & \textbf{0.98} & 0.97 & \textbf{0.98} & \textbf{0.98} & \textbf{0.98} & 0.91 & \textbf{0.98} & 0.91 \\
 & Time & 7:57:41 & \textbf{1:10:35} & 9:39:52 & 1:27:02 & 10:04:43 & 1:26:28 & 10:43:40 & 1:25:19 \\
\hline
\multirow{3}{*}{blog}
 & \% Reduction & 49.79 & \textbf{80.50} & 34.29 & 72.86 & 34.79 & 75.21 & 35.79 & 74.14 \\
 & Test Score & 0.37 & 0.35 & \textbf{0.41} & 0.37 & 0.39 & 0.30 & 0.38 & 0.38 \\
 & Time & 0:11:33 & \textbf{0:09:04} & 0:13:56 & 0:10:31 & 0:14:08 & 0:11:36 & 0:13:53 & 0:10:35 \\
\hline
\multirow{3}{*}{bss$^\dagger$}
 & \% Reduction & 53.37 & \textbf{79.71} & 54.18 & 69.08 & 54.58 & 73.94 & 58.80 & 74.51 \\
 & Test Score & \textbf{0.79} & 0.78 & \textbf{0.79} & 0.78 & \textbf{0.79} & 0.75 & \textbf{0.79} & 0.74 \\
 & Time & 6:01:17 & \textbf{1:33:37} & 7:35:38 & 2:27:30 & 7:49:42 & 2:10:02 & 7:58:37 & 2:05:07 \\
\hline
\multirow{3}{*}{carer$^\dagger$}
 & \% Reduction & 47.14 & \textbf{67.76} & 37.14 & 54.90 & 36.35 & 60.21 & 38.54 & 62.19 \\
 & Test Score & 0.62 & 0.61 & \textbf{0.63} & 0.62 & \textbf{0.63} & 0.61 & \textbf{0.63} & 0.61 \\
 & Time & 4:29:03 & \textbf{2:47:45} & 6:00:23 & 3:21:13 & 6:14:05 & 3:50:33 & 6:03:29 & 3:10:43 \\
\hline
\multirow{3}{*}{colon}
 & \% Reduction & 56.3 & \textbf{83.76} & 51.87 & 83.1 & 36.86 & 77.16 & 48.32 & 78.50 \\
 & Test Score & 0.88 & \textbf{0.92} & 0.88 & 0.85 & 0.80 & 0.84 & 0.85 & 0.84 \\
 & Time & 0:02:43 & \textbf{0:01:44} & 0:02:56 & 0:01:50 & 0:03:12 & 0:02:12 & 0:03:12 & 0:01:54 \\
\hline
\multirow{3}{*}{cyber$^\dagger$}
 & \% Reduction & 51.67 & \textbf{73.75} & 51.69 & 65.73 & 45.73 & 68.70 & 44.69 & 69.22 \\
 & Test Score & \textbf{0.79} & 0.78 & \textbf{0.79} & 0.78 & \textbf{0.79} & 0.77 & \textbf{0.79} & 0.77 \\
 & Time & 4:23:54 & \textbf{2:23:32} & 5:41:04 & 2:42:03 & 6:00:34 & 3:02:48 & 5:58:06 & 2:36:53 \\
\hline
\multirow{3}{*}{gisette}
 & \% Reduction & 64.01 & \textbf{95.89} & 77.14 & 94.46 & 68.38 & 94.82 & 66.66 & 95.13 \\
 & Test Score & \textbf{1.00} & \textbf{1.00} & \textbf{1.00} & \textbf{1.00} & 1.00 & 0.98 & 1.00 & 0.98 \\
 & Time & 4:02:02 & \textbf{0:20:11} & 4:47:05 & 0:21:32 & 5:01:58 & 0:22:13 & 5:21:43 & 0:21:16 \\
\hline
\multirow{3}{*}{house}
 & \% Reduction & 55.44 & \textbf{77.72} & 51.90 & 72.91 & 46.33 & 73.16 & 45.06 & 74.68 \\
 & Test Score & 0.84 & 0.80 & 0.82 & 0.81 & 0.82 & 0.69 & 0.84 & 0.72 \\
 & Time & 0:05:07 & \textbf{0:04:32} & 0:05:32 & 0:04:47 & 0:05:37 & 0:05:24 & 0:05:42 & 0:04:48 \\
\hline
\multirow{3}{*}{isolet}
 & \% Reduction & 57.18 & \textbf{68.40} & 46.55 & 57.83 & 38.06 & 60.13 & 39.84 & 60.91 \\
 & Test Score & 0.66 & 0.66 & \textbf{0.67} & 0.66 & \textbf{0.67} & 0.57 & \textbf{0.67} & 0.57 \\
 & Time & 1:05:43 & \textbf{0:38:17} & 1:27:42 & 0:52:31 & 1:29:52 & 0:51:41 & 1:27:34 & 0:50:41 \\
\hline
\multirow{3}{*}{madelon}
 & \% Reduction & 59.08 & \textbf{82.24} & 56.30 & 82.00 & 47.84 & 77.12 & 40.12 & 77.40 \\
 & Test Score & \textbf{0.87} & \textbf{0.87} & \textbf{0.87} & 0.86 & 0.86 & 0.65 & \textbf{0.87} & 0.62 \\
 & Time & 0:14:53 & \textbf{0:08:03} & 0:17:50 & 0:09:04 & 0:18:02 & 0:10:14 & 0:18:09 & 0:09:27 \\
\hline
\multirow{3}{*}{mnist}
 & \% Reduction & 60.71 & \textbf{76.48} & 65.59 & 65.33 & 66.15 & 70.18 & 65.28 & 71.28 \\
 & Test Score & \textbf{0.88} & 0.87 & \textbf{0.88} & 0.87 & \textbf{0.88} & 0.83 & \textbf{0.88} & 0.83 \\
 & Time & 10:21:10 & \textbf{5:51:23} & 13:20:12 & 7:05:44 & 13:29:41 & 7:26:11 & 14:29:39 & 6:49:41 \\
\hline
\multirow{3}{*}{relathe}
 & \% Reduction & 56.68 & \textbf{93.21} & 53.73 & 90.44 & 51.19 & 91.59 & 51.62 & 91.57 \\
 & Test Score & \textbf{0.95} & \textbf{0.95} & \textbf{0.95} & \textbf{0.95} & \textbf{0.95} & 0.85 & \textbf{0.95} & 0.89 \\
 & Time & 0:15:19 & \textbf{0:04:59} & 0:18:47 & \textbf{0:04:59} & 0:19:29 & 0:05:27 & 0:19:55 & \textbf{0:04:59} \\
\hline
\multirow{3}{*}{spam$^\dagger$}
 & \% Reduction & 63.38 & \textbf{83.17} & 66.81 & 78.39 & 64.42 & 79.95 & 63.06 & 81.45 \\
 & Test Score & \textbf{0.98} & \textbf{0.98} & \textbf{0.98} & \textbf{0.98} & \textbf{0.98} & 0.97 & \textbf{0.98} & 0.97 \\
 & Time & 0:50:45 & \textbf{0:16:43} & 1:02:00 & 0:19:37 & 1:03:24 & 0:22:10 & 1:08:17 & 0:18:46 \\
\hline
\multirow{3}{*}{spam\_meta\_embed$^\dagger$}
 & \% Reduction & 62.70 & \textbf{94.47} & 75.14 & 93.58 & 68.30 & 92.93 & 68.55 & 93.35 \\
 & Test Score & \textbf{0.99} & \textbf{0.99} & \textbf{0.99} & \textbf{0.99} & \textbf{0.99} & \textbf{0.99} & \textbf{0.99} & \textbf{0.99} \\
 & Time & 1:23:50 & 0:12:09 & 1:45:43 & \textbf{0:11:58} & 1:49:12 & 0:14:07 & 1:57:00 & 0:12:32 \\
\hline
\end{tabular}
}
\\[1.2ex]
\footnotesize{$^\dagger$ Datasets containing text samples, with features corresponding to sentence embedding dimensions computed using an LLM.}
\end{center}
\label{t:full_results_comparison_hybrid}
\end{table*}

\end{document}